\documentclass{article}

\usepackage{microtype}
\usepackage{graphicx}
\usepackage{subcaption}
\usepackage{booktabs}
\usepackage{multirow}
\usepackage{float}
\usepackage{placeins}

\usepackage{hyperref}

\usepackage[accepted]{icml2026}

\makeatletter
\renewcommand{\ICML@appearing}{%
  \textit{ICML 2026 Workshop on Resource-Adaptive Foundation Model Inference},
  Seoul, South Korea. Copyright 2026 by the author(s).%
}
\renewcommand{\Notice@String}{\ICML@appearing}
\makeatother

\usepackage{amsmath}
\usepackage{amssymb}
\usepackage{mathtools}
\usepackage{amsthm}

\usepackage[capitalize,noabbrev]{cleveref}

\newcommand{\dPPL}{\Delta\text{PPL}}
\newcommand{\dhead}{d_h}
\newcommand{\softmax}{\operatorname{softmax}}
\newcommand{\sig}{\sigma}

\theoremstyle{plain}

\icmltitlerunning{Sigmoid Attention as a Better Substrate for Learned KV Cache Eviction}

\hypersetup{
  pdftitle={Sigmoid Attention as a Better Substrate for Learned KV Cache Eviction},
  pdfauthor={Isaac (Rucheng) Li},
  pdfsubject={ICML 2026 Workshop on Resource-Adaptive Foundation Model Inference},
  pdfkeywords={KV cache compression, sigmoid attention, efficient inference, resource-adaptive inference, learned eviction}
}

\begin{document}

\twocolumn[
  \icmltitle{Sigmoid Attention as a Better Substrate for\\
             Learned KV Cache Eviction}

  \begin{icmlauthorlist}
  \icmlauthor{Isaac (Rucheng) Li}{pitt}
  \end{icmlauthorlist}

  \icmlaffiliation{pitt}{University of Pittsburgh, Pittsburgh, PA, USA}

  \icmlcorrespondingauthor{Isaac (Rucheng) Li}{isl74@pitt.edu}

  \icmlkeywords{KV cache compression, sigmoid attention, efficient inference,
                resource-adaptive inference, learned eviction}

  \vskip 0.3in
]

\printAffiliationsAndNotice{}

\begin{abstract}
Learned KV-cache eviction often faces a soft-to-hard mismatch: during training, differentiable gates typically attenuate token contributions, whereas inference saves memory only when KV entries are physically removed. We ask whether the attention substrate affects this
soft-to-hard transition. Using GPT-2-scale Transformers trained on
OpenWebText, we run a controlled $2\times2\times2$ comparison over
attention type, learned gating, and positional encoding. Although
sigmoid attention is worse as a dense language model, learned hard
eviction changes the useful operating points: sigmoid-gated models
delete KV entries with negligible PPL change relative to their own
no-eviction references. Under a matched live-cache protocol on the same dense backbones,
learned sigmoid gates obtain lower PPL than our H$_2$O and KeyDiff
implementations, whereas softmax gates do not uniformly beat these
post-hoc methods.
The results suggest that attention normalization can substantially affect
whether a training-time soft gate transfers cleanly to hard KV deletion.
\end{abstract}

\section{Introduction}
\label{sec:intro}

Learned KV-cache eviction has a soft-to-hard mismatch. During training,
a model can only learn a differentiable proxy for deletion: a token is
attenuated by a soft gate, but its key and value remain in memory. During
inference, however, memory is freed only when the corresponding KV entry
is physically removed. A useful learned eviction policy must therefore
be threshold-stable: replacing soft attenuation by hard deletion
at some threshold $\tau$ should introduce little or no loss in
next-token prediction quality.

Most KV-cache compression work does not directly ask which attention
substrate makes this soft-to-hard transition easier to learn. Online
post-hoc methods operate on a frozen model and select which cache entries
to retain using signals available at inference time. H$_2$O maintains
accumulated attention-based heavy-hitter scores and retains high-score
tokens together with recent tokens \citep{zhang2023h2o}. KeyDiff is an
attention-free, key-similarity-based method that ranks cached keys by
their similarity to a key-space anchor and retains distinctive keys
\citep{keydiff}. Learned compression methods, including Dynamic Context
Pruning and Dynamic Memory Compression, show that retention policies can
be trained \citep{anagnostidis2023dynamic,nawrot2024dmc}. Our question
is complementary: holding the learned eviction mechanism fixed, does the
attention normalization itself affect how learnable hard KV deletion is?

We test this question with a controlled $2\times2\times2$ experiment:
softmax versus sigmoid attention, dense versus learned eviction, and
RoPE versus NoPE. NoPE denotes the removal of rotary embeddings from queries and keys. All models share the same GPT-2-scale setup. Gated
models use the same soft-to-hard rule: token gates attenuate logits and
values during training, then delete KV entries with $g<\tau$ at inference.

The results separate dense modeling quality from eviction learnability.
Without eviction, softmax remains stronger: sigmoid attention is worse
by $\sim 0.36$ PPL with RoPE and $\sim 1.21$ PPL without RoPE. With
learned hard eviction, however, the useful operating points come from
sigmoid attention. Against their own dense no-eviction references,
Sig+G RoPE deletes $19.8\%$ of KV entries with no measurable PPL penalty
($22.424$ vs.\ $22.440$), while Sig+G NoPE deletes $32.2\%$ with only a
small increase ($24.637$ vs.\ $24.603$). Under the same matched live-cache protocol, both learned sigmoid gates
obtain lower PPL than our H$_2$O and KeyDiff implementations. By contrast, softmax-gated models lose to H$_2$O at matched
compression.

Further analysis clarifies why this pattern is substrate-specific. A
random support-removal probe shows that sigmoid attention is not
generically more robust to arbitrary K/V deletion: when random K/V
positions are removed in dense RoPE models, sigmoid attention outputs
are more perturbed than softmax at larger removal rates. Thus the
advantage does not come from sigmoid tolerating deletion
indiscriminately. It appears when the deletion pattern is learned,
where non-row-normalized attention gives soft gates a cleaner local attenuation channel.

Code and notebooks for reproducing the eight experimental cells are available at \url{https://github.com/IsaacLi74/sigmoid-kv-eviction}.

\section{Method}
\label{sec:method}

\subsection{Factorial model family}
We evaluate eight models in a $2\times2\times2$ design over
\textit{attention} (softmax / sigmoid), \textit{gate} (dense / learned),
and rotary position treatment (RoPE~\citep{su2021rope} / NoPE). In this
paper, NoPE denotes the no-RoPE condition used in our implementation:
the RoPE rotation is removed from queries and keys. It does \emph{not}
use GPT-2-style learned absolute position embeddings; the input
representation is the tied token embedding only,
\[
x_i^{(0)} = W_E[t_i],
\]
with no added learned position vector. All cells share a
GPT-2--style decoder-only Transformer
\citep{vaswani2017attention,radford2019gpt2} with
$n_{\text{layer}}{=}12$, $n_{\text{head}}{=}12$,
$d_{\text{model}}{=}768$, $\dhead{=}64$, $d_{\text{ff}}{=}3072$, and
sequence length $512$. Training uses AdamW~\citep{loshchilov2019adamw} for three epochs on a
$1\mathrm{B}$-token OpenWebText training split
\citep{gokaslan2019owt}, corresponding to roughly $3\mathrm{B}$ token
exposures. We use held-out splits of $100\mathrm{K}$ tokens for
validation and $1\mathrm{M}$ tokens for test.

\subsection{Attention variants}

All variants apply per-head QK-RMSNorm~\citep{zhang2019rmsnorm} before
computing attention scores:
\[
\tilde q_i = \mathrm{RMSNorm}(q_i), \qquad
\tilde k_j = \mathrm{RMSNorm}(k_j).
\]

In RoPE cells, we then rotate queries and keys with RoPE; in NoPE cells,
this rotation is omitted:
\[
(\bar q_i, \bar k_j) =
\begin{cases}
(\mathrm{RoPE}(\tilde q_i,i), \mathrm{RoPE}(\tilde k_j,j)), & \text{RoPE},\\
(\tilde q_i,\tilde k_j), & \text{NoPE}.
\end{cases}
\]

\paragraph{Softmax.}
\[
A_{ij}
=
\softmax_j\!\left(
\bar q_i^\top \bar k_j / \sqrt{\dhead}
\right).
\]

\paragraph{Sigmoid attention.}
\begin{equation*}
A_{ij} = \sig\!\left( \bar q_i^\top \bar k_j / \sqrt{\dhead} + b(i)\right),
\quad b(i) = -\log(i+1).
\end{equation*}
Following \citet{ramapuram2024sigmoid}, we include a negative bias term
to stabilize sigmoid attention norms. One practical choice discussed by \citet{ramapuram2024sigmoid} is a
negative scalar bias on the order of \(-\log n\), where $n$ is the maximum training sequence
length. Since a causal query at position \(i\) attends to only \(i+1\)
keys, we use the per-query bias \(b(i)=-\log(i+1)\) to stabilize the
expected row mass at each position. This row-wise bias would be a
mathematical no-op under softmax, since adding the same constant to all
logits in a row leaves the softmax distribution unchanged.

\subsection{Learned gate}
For gated models, each layer $\ell<L{-}1$ produces a token-wise gate
\begin{equation*}
g_j^{(\ell)} = \sig\!\left( w_\ell^\top x_j^{(\ell)} + b_\ell \right) \in (0,1)
\end{equation*}
which controls how the next layer sees token~$j$. Let
\(z_{ij}^{(\ell+1)}\) denote the pre-attention logit in layer \(\ell+1\).
We inject the gate through both the attention logits and the values:
\begin{equation*}
z_{ij}^{(\ell+1)}
\leftarrow
z_{ij}^{(\ell+1)}
+
\log\!\left(g_j^{(\ell)}+\varepsilon\right),
\qquad
v_j^{(\ell+1)}
\leftarrow
g_j^{(\ell)} v_j^{(\ell+1)}.
\end{equation*}
where $\varepsilon=10^{-8}$ avoids $\log 0$. This is the \emph{same} injection for both attention types; the last
layer carries no gate (no subsequent KV cache to evict). Gates are initialized with $w_\ell = 0$ and $b_\ell = 5$, so $g_j^{(\ell)} =
\sig(5) \approx 0.993$ at step 0 and early training behaves like the
no-gate model.

The two-channel injection ($\log g$ on logits, $g$ on values) is a
differentiable \emph{simulation} of dropping: as $g\to 0$, both channels
drive token $j$'s contribution to zero, matching the limiting behavior
of hard deletion. In inference, we convert this soft retention signal into physical cache
removal by dropping entries with \(g_j < \tau\). Entries removed after
prefill are unavailable to all subsequent decode steps; during decode,
the same threshold rule is applied to each newly written entry. Entries
with \(g_j\ge\tau\) remain in the cache and continue to use the same
learned logit and value scaling.

\subsection{Objective}
We use a simple training objective: cross-entropy plus a mean budget term
that pulls gates toward small values,
\begin{equation*}
  \mathcal{L} \;=\; \mathcal{L}_{\mathrm{CE}} \;+\; \lambda\,\overline{g},
  \qquad
  \overline{g} \;=\; \frac{1}{(L-1)\,T}
    \sum_{\ell=0}^{L-2}\sum_{j=1}^{T} g_j^{(\ell)},
\end{equation*}
with $\lambda=0.03$. We do not tune \(\lambda\) per cell; the same budget coefficient is used for all gated variants.

\subsection{Evaluation}
\label{sec:eval}
\paragraph{Test protocol.} On the OpenWebText test split, each
sequence is split into a prefix of $384$ tokens for prefill and a window
of $128$ tokens for single-token decode; PPL is computed on the decode window.

\paragraph{Threshold selection.} For each gated model we sweep a threshold grid from \(0\) to \(0.9\) on
$100$ sequences from the held-out validation split and select the
highest-compression $\tau^\star$ satisfying $\dPPL<0.1$ versus that gated model's no-eviction reference. The selected $\tau^\star$ is then frozen
for test.

\paragraph{Physical eviction.} Learned-gate evaluation performs
thresholded physical eviction: after prefill we remove every cached entry
with \(g_j<\tau^\star\), and during decode the same rule is applied to each
newly written entry. Entries that remain in the cache still use the
learned logit and value scaling from \cref{sec:method}. Reported
compression rates count only physically removed KV entries, not the
additional soft attenuation applied to retained entries.

\paragraph{Matched live-cache size.}
For a sequence, the live-cache size is the number of KV entries that
remain after hard eviction. Compression is measured relative to the
corresponding no-eviction cache. When comparing learned gates with
post-hoc baselines, we first measure the learned gate's realized final
live-cache size, then run post-hoc methods with the same final
live-cache size on the paired dense no-gate backbone. Thus the baselines need
not delete the same tokens, but they operate at the same cache size.

\paragraph{Online post-hoc baselines.}
On each dense backbone we run two online post-hoc KV-cache eviction
baselines at the matched final live-cache size. H$_2$O is an
attention-score-based heavy-hitter method: it maintains layer-wise,
per-head cumulative attention scores and retains a mixture of high-score
historical tokens and recent tokens \citep{zhang2023h2o}. For H$_2$O,
we split the matched cache size equally between heavy-hitter and recent
tokens and do not tune this ratio on validation. KeyDiff is an
attention-free baseline: for each layer and head, it computes an
unnormalized cached-key anchor $\mu(K)$, ranks cached entries by cosine
similarity to this anchor, and retains the least similar keys
\citep{keydiff}. Both baselines are run on the corresponding dense
backbone, not on the gated checkpoint. This comparison isolates learned
soft-to-hard eviction against post-hoc eviction under the same
attention--RoPE substrate.

\section{Results}
\label{sec:results}

We report results in three steps. First, dense no-eviction PPL separates
language-modeling quality from eviction behavior
(\cref{sec:no_eviction_baseline}). Second, validation threshold sweeps
select one frozen hard-deletion operating point per gated model
(\cref{sec:tau_selection}). Third, matched final live-cache size test
comparisons evaluate whether learned gates improve over post-hoc
eviction on the corresponding dense backbones. Threshold selection uses
each gated checkpoint at $\tau=0$ as the no-eviction reference. In the
matched live-cache comparison, post-hoc baselines are run on the paired
dense no-gate backbone, so we also report the learned gate's change
relative to that dense backbone.

\subsection{Dense baselines: softmax has lower no-eviction PPL}
\label{sec:no_eviction_baseline}

\cref{tab:no_eviction} reports dense no-eviction decode-window PPL for
the four attention--RoPE backbones under the same prefill--decode
protocol.

\begin{table}[h]
\centering
\caption{Dense no-eviction test PPL.}
\label{tab:no_eviction}
\begin{tabular}{lc}
\toprule
Backbone & Dense PPL \\
\midrule
SM RoPE   & 22.079 \\
Sig RoPE  & 22.440 \\
SM NoPE   & 23.398 \\
Sig NoPE  & 24.603 \\
\bottomrule
\end{tabular}
\end{table}

Sigmoid attention is not a stronger dense language model in this setup:
softmax is better by $0.361$ PPL with RoPE and $1.205$ PPL without RoPE.

\subsection{Validation threshold selection}
\label{sec:tau_selection}

\begin{figure}[H]
\centering
\includegraphics[width=\linewidth]{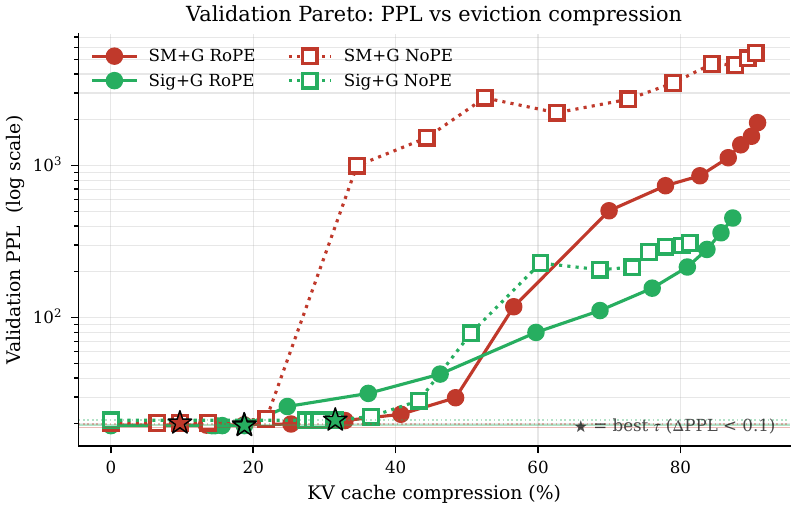}
\caption{Validation threshold sweep. Each point shows the validation PPL
and realized hard-KV compression rate for one threshold $\tau$. Stars mark
the selected $\tau^\star$ for each gated model.}
\label{fig:val_sweep}
\end{figure}

For each gated model we sweep the hard-deletion threshold over a
15-point validation grid and select the highest-compression threshold
satisfying $\dPPL<0.1$ against that model's $\tau=0$ reference. The
selected threshold is frozen before test evaluation. \cref{fig:val_sweep}
shows how validation PPL changes as compression increases under the
threshold sweep, and \cref{tab:tau_selection} summarizes the selected
operating points. The full numerical sweep is reported in
\cref{tab:app_tau_sweep}.

The threshold curves already separate the substrates. Both softmax
models select $\tau^\star=0.01$, while both sigmoid models select
$\tau^\star=0.05$. With RoPE, softmax and sigmoid reach nearly matched
compression, but Sig+G has essentially no validation PPL increase.
Without RoPE, Sig+G reaches $31.5\%$ compression, compared with $9.7\%$
for SM+G.

\begin{table}[H]
\centering
\caption{Validation-selected operating points. Compression is hard
KV-entry deletion during prefill and decode.}
\label{tab:tau_selection}
\begin{tabular}{lccc}
\toprule
Model & $\tau^\star$ & $\dPPL$ & Compression \\
\midrule
SM+G RoPE  & 0.010 & $+0.077$ & $18.7\%$ \\
Sig+G RoPE & 0.050 & $-0.001$ & $18.8\%$ \\
SM+G NoPE  & 0.010 & $+0.016$ & $\phantom{0}9.7\%$ \\
Sig+G NoPE & 0.050 & $+0.033$ & $31.5\%$ \\
\bottomrule
\end{tabular}
\end{table}

\subsection{Matched live-cache size results}
\label{sec:test_operating}

On test, we freeze the validation-selected $\tau^\star$ and compare each
learned gate to post-hoc H$_2$O and KeyDiff run on the corresponding
dense backbone at the matched final live-cache size.

\begin{table}[h]
\centering
\small
\caption{Frozen-threshold test operating points. H$_2$O and KeyDiff are run on the corresponding dense backbone at the
same final live-cache size as the learned gate.}
\label{tab:test_headline}
\resizebox{\columnwidth}{!}{%
\begin{tabular}{lccccc}
\toprule
Model & $\tau^\star$ & Comp. & Learned PPL & H$_2$O & KeyDiff \\
\midrule
SM+G RoPE
  & 0.010 & $19.2\%$ & 22.514 & \textbf{22.145} & 22.984 \\
Sig+G RoPE
  & 0.050 & $19.8\%$ & \textbf{22.424} & 22.480 & 22.474 \\
SM+G NoPE
  & 0.010 & $10.2\%$ & 23.576 & \textbf{23.423} & 23.588 \\
Sig+G NoPE
  & 0.050 & $32.2\%$ & \textbf{24.637} & 25.107 & 25.786 \\
\bottomrule
\end{tabular}%
}
\end{table}

The test results show that learned gates are not uniformly better than
post-hoc eviction. On softmax backbones, H$_2$O is stronger than the
learned gate at the matched final live-cache size. On sigmoid backbones, the pattern reverses under the same matched-cache
protocol. Sig+G RoPE deletes $19.8\%$ of KV entries and slightly improves
the measured PPL relative to its dense no-eviction reference
($22.424$ vs.\ $22.440$), while obtaining lower PPL than both post-hoc
baselines at the same final live-cache size. Sig+G NoPE deletes $32.2\%$
with only a small increase over its dense reference
($24.637$ vs.\ $24.603$) and again obtains lower PPL than the matched
H$_2$O and KeyDiff runs.

Thus the main result is substrate-specific. The same simple learned
soft-to-hard rule is weak under softmax, but becomes useful
under sigmoid attention. The RoPE and NoPE regimes show different forms
of this effect: with RoPE, sigmoid mainly improves quality at nearly
matched compression; without RoPE, it mainly expands the safe
compression range.

\FloatBarrier

\section{Mechanism Analysis}
\label{sec:analysis}

The matched live-cache results suggest that the advantage is not a
generic property of sigmoid attention, but a property of learned deletion
under sigmoid attention. We use three diagnostics to support this
interpretation: random support removal, row-sum and gate statistics, and
token-level selectivity.

\subsection{Not arbitrary-deletion robustness}
\label{sec:not_generic}

A possible alternative explanation is that sigmoid attention simply
tolerates arbitrary K/V removal better than softmax. We test this with a
random support-removal probe on the two dense RoPE backbones, with no
learned gates involved. For each sampled test sequence and each layer,
we first record the normal attention output
\[
O_{s,\ell}^{\mathrm{full}} = A_{s,\ell}^{\mathrm{full}}V_{s,\ell}^{\mathrm{full}} .
\]
We then run a second pass in which a random fraction $p$ of K/V positions
is removed from the attention support at every layer. The first four
positions are excluded to avoid confounding from attention-sink artifacts
\citep{xiao2024streamingllm,gu2025attention}. Removed positions
are masked out of the attention logits and their value vectors are
zeroed, so they cannot receive attention mass or contribute content.

\begin{figure}[H]
\centering
\includegraphics[width=\linewidth]{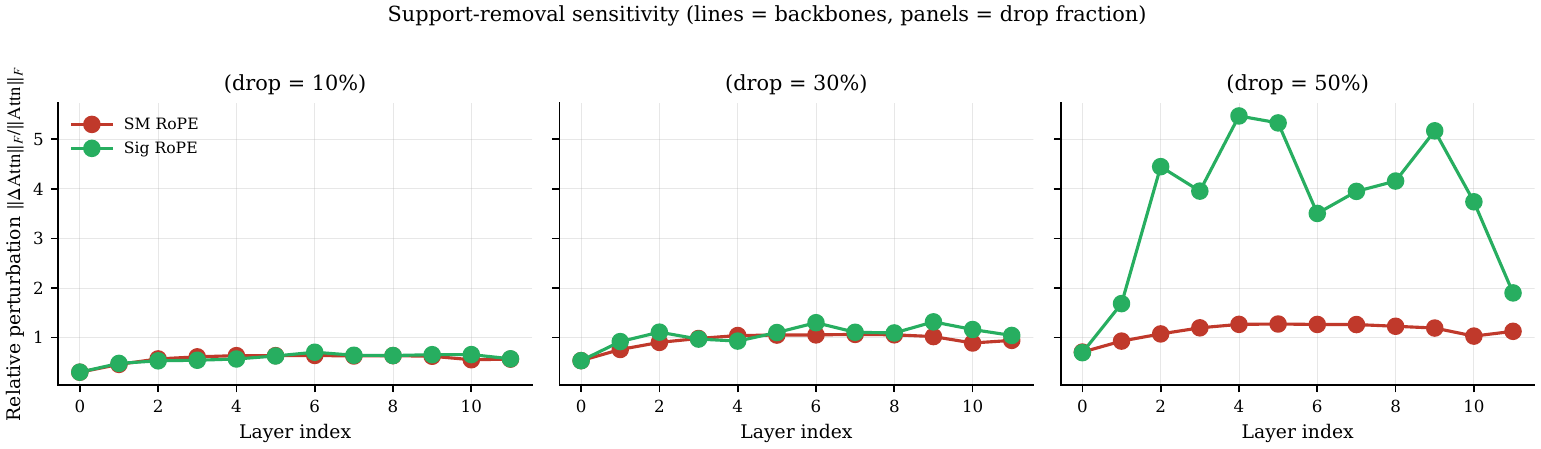}
\caption{Random support-removal sensitivity on dense RoPE models. For
each drop rate $p$, we randomly remove K/V support and measure the
relative perturbation of each layer's attention output.}
\label{fig:support_removal}
\end{figure}

We measure the resulting layerwise perturbation by
\[
\rho_{s,\ell}(p)
=
\frac{
\left\|O_{s,\ell}^{\mathrm{drop}}(p)
-
O_{s,\ell}^{\mathrm{full}}\right\|_F
}{
\left\|O_{s,\ell}^{\mathrm{full}}\right\|_F
}.
\]
For each model and drop rate, we sample one random mask per sequence and
per layer. \cref{fig:support_removal} plots the resulting perturbation
for each layer after averaging over $50$ test sequences. This probe is a
representation-level sensitivity test, not a random-eviction PPL
baseline.

The probe does not support generic deletion robustness. To summarize the
curves, we average the sequence-averaged perturbations over all $12$
layers. At $p{=}0.1$, softmax and sigmoid are nearly tied
($0.574$ vs.\ $0.579$). At larger removal rates, sigmoid is more
perturbed: $0.943$ vs.\ $1.051$ at $p{=}0.3$, and $1.131$ vs.\ $3.666$
at $p{=}0.5$. Thus the sigmoid operating-point advantage is unlikely to
come from tolerating arbitrary K/V removal. It appears when the deletion
pattern is learned.

\subsection{Row-sum freedom and gate stability}
\label{sec:rowsum_freedom}

The next diagnostic asks why learned soft gates may transfer more
cleanly to hard deletion under sigmoid attention. The key structural
difference is row normalization. In softmax attention, each row sums to
one, so suppressing one token redistributes attention mass to the other
tokens in the same row. Thus a gate changes not only the gated token's
contribution, but also the relative weights of the remaining tokens.

Sigmoid attention does not impose this constraint. Its row mass
$\sum_j a_{ij}$ can vary with the query, layer, and sequence. A small
gate can therefore mainly attenuate the selected token's own contribution
without forcing the rest of the row to compensate. This can make the
training-time soft operation a closer approximation to hard deletion for
the gated token, because the remaining row is not forced to renormalize.

We measure this row-sum freedom using the coefficient of variation
\[
\mathrm{CV}
=
\frac{
\mathrm{std}_i\left(\sum_j a_{ij}\right)
}{
\mathrm{mean}_i\left(\sum_j a_{ij}\right)
},
\]
computed across query positions on $1953$ test sequences of length
$512$. For softmax, this value is zero by construction because every row
sums to one. For sigmoid, a nonzero value means that different query
positions can carry different total attention mass. We also track gate
binarization, defined as the fraction of gates with $g<0.05$ or
$g>0.95$, as a measure of whether training forms a thresholdable
keep/delete structure.

\begin{figure}[h]
\centering
\includegraphics[width=\linewidth]{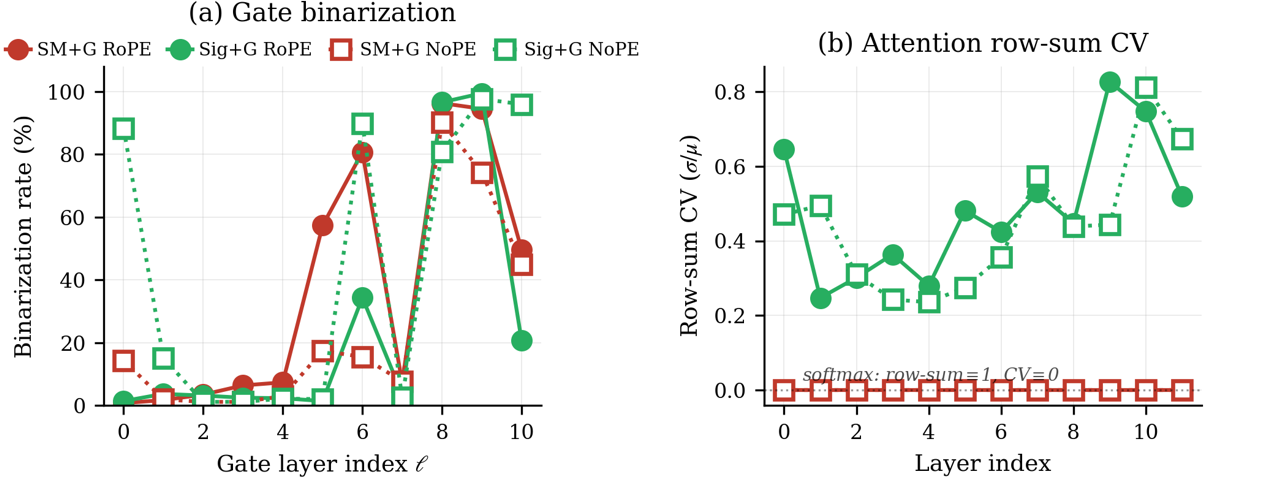}
\caption{Per-layer mechanism evidence. Left: gate binarization. Right:
coefficient of variation of attention row sums.}
\label{fig:mechanism}
\end{figure}

\cref{fig:mechanism} shows these two ingredients separately. The right
panel confirms the structural difference: softmax has zero row-sum
variation by construction, while sigmoid retains nonzero row-sum
variation across layers. Averaged over layers, the row-sum CV is
$0.484$ for Sig+G RoPE and $0.444$ for Sig+G NoPE, compared with $0$
for both softmax models. The left panel shows that learned gates can
also form near-binary structure, especially in the high-compression
Sig+G NoPE model.

Averaged across gate layers and test token positions, Sig+G NoPE has
the strongest aggregate binarization ($43.3\%$), while SM+G RoPE is also
substantially binarized ($36.8\%$). However, binarization alone does not
explain the result: Sig+G RoPE has lower aggregate binarization than
SM+G RoPE ($24.4\%$ vs.\ $36.8\%$), yet transfers more cleanly to hard
deletion. The useful regime therefore appears to require both pieces:
the gates must learn a thresholdable structure, and the attention
substrate must let soft attenuation approximate hard deletion locally.

\subsection{Token selectivity}
\label{sec:token_selectivity}

The final diagnostic asks what the learned gates choose to suppress. If
the gates were merely uniform cache shrinkers, then all token types would
receive similar gate values. Instead, the gates are token-selective.

We bucket GPT-2 surface tokens into coarse categories: content words,
stopwords, punctuation, newline, whitespace, and digits. For each bucket,
\cref{tab:token_buckets} reports the mean learned gate value on the test
split. Lower values mean the model is more willing to remove that kind
of token under hard thresholding; higher values mean the model tends to
preserve it.

\begin{table}[H]
\centering
\small
\caption{Mean gate value by token type on the test split. Buckets are
assigned from the GPT-2 decoded surface form.}
\label{tab:token_buckets}
\setlength{\tabcolsep}{3pt}
\resizebox{\columnwidth}{!}{%
\begin{tabular}{lcccccc}
\toprule
Model & content & stopword & punct. & newline & whitesp. & digit \\
\midrule
SM+G RoPE  & 0.213 & 0.144 & 0.224 & 0.170 & 0.274 & 0.196 \\
Sig+G RoPE & 0.272 & 0.184 & 0.270 & 0.176 & 0.328 & 0.235 \\
SM+G NoPE  & 0.223 & 0.144 & 0.195 & 0.135 & 0.334 & 0.224 \\
Sig+G NoPE & 0.281 & 0.178 & 0.254 & 0.273 & 0.455 & 0.263 \\
\bottomrule
\end{tabular}%
}
\end{table}

The pattern is consistent across all four gated models. Across models, stopwords and newlines tend to receive the lowest gates,
while whitespace tokens consistently receive the highest gates. Digit
tokens are generally assigned higher gates than stopwords, but are not
always among the highest-retention categories. For example, Sig+G NoPE assigns stopwords a mean gate of
$0.178$, compared with $0.281$ for content tokens and $0.455$ for
whitespace tokens. The token-level rankings show the same qualitative pattern. Soft-gate
rankings frequently place common function words such as \verb| the|,
\verb| a|, \verb| an|, \verb|The|, and \verb|It| among low-gate tokens,
while the actual hard-threshold eviction rankings include frequent
function words such as \verb| the|, \verb| a|, \verb| an|, \verb| of|,
\verb| in|, \verb| on|, \verb| for|, \verb| to|, and \verb| be|. This means the learned policy is
not deleting a fixed fraction of the cache uniformly; it assigns
different retention values to different token classes.

This also rules out a simple ``smaller gates are better'' explanation.
Sigmoid models assign higher mean gates than softmax models in every
bucket, so the sigmoid advantage does not come from more aggressive soft
suppression during training. The advantage appears when those gates are
hard-thresholded: under the sigmoid substrate, the learned gate values
form a more useful keep/delete partition. Qualitative gate heatmaps on
five short prompts are shown in \cref{app:gate_heatmaps}; they illustrate
the same non-uniform structure across layers, token positions, and
substrates.

Taken together, these diagnostics support a substrate-level finding.
Sigmoid attention is not generically robust to arbitrary K/V removal:
random support removal perturbs it more at larger removal rates. Its
advantage appears when deletion is learned. Because sigmoid attention is
not row-normalized, soft attenuation of a token can be closer to hard
deletion of that token than under a row-normalized softmax substrate. The learned gates then select token classes
non-uniformly rather than shrinking the cache by a fixed ratio.
\FloatBarrier

\section{Discussion}
\label{sec:discussion}

\subsection{Interpretation}
\label{sec:discussion_interpretation}

The main implication of our results is not that sigmoid attention is a
strictly better language-modeling substrate. In the dense no-eviction
setting, softmax remains stronger. Rather, sigmoid attention appears to
be a better substrate for a specific systems objective: training a soft
retention signal that can later be converted into hard physical KV
deletion. This distinction is important. Post-hoc eviction methods test
how well a frozen model tolerates externally imposed cache removal. Our
experiment instead asks whether the attention substrate makes a learned
eviction policy easier to train in the first place.

This view also changes how to interpret the strongest operating point.
Sig+G RoPE physically deletes $19.8\%$ of KV entries while achieving
slightly lower measured PPL than its dense no-eviction backbone
($22.424$ vs.\ $22.440$). We do not interpret this small negative
$\dPPL$ as a reliable language-modeling gain. Its significance is more
operational: learned eviction can remove a nontrivial fraction of the
cache without measurable quality loss, and may sometimes act as
selective cache denoising rather than merely lossy compression.

\subsection{Implications}
\label{sec:discussion_implications}

If this substrate effect persists at larger scale and longer context length, it would motivate joint design of attention mechanisms and KV-retention policies. Rather than treating KV eviction only as an inference-time heuristic, our results frame it as an architectural co-design problem.

\subsection{Limitations}
\label{sec:discussion_limitations}

Our study has several limitations. First, we pick a single
$\tau^\star$ per gated model from a discrete validation grid by
maximizing compression subject to $\dPPL{<}0.1$. This rule is simple and
frozen before test evaluation, but the cutoff is still a design choice.
Finer threshold grids, per-layer thresholds, or alternative operating
criteria could select different cache sizes. We also compare post-hoc
baselines against learned gates at matched final live-cache size rather
than matched PPL. This directly controls memory use, which is often the deployment
constraint of interest, but it does not answer every deployment question;
sigmoid threshold sweeps can have sharp cliffs, making matched-PPL
comparison less stable.

Second, experiments use one $123\mathrm{M}$ GPT-2--scale model family
trained for three epochs on a $1\mathrm{B}$-token OpenWebText split, with a single seed per cell and
context length $512$. This is sufficient for a controlled substrate
study, but not sufficient to establish deployment-scale behavior.
Whether the effect persists at larger model scales, longer contexts,
multiple seeds, other text distributions, or downstream long-context
tasks remains open.

Third, we use one gate design: token-wise gates with both logit
attenuation and value scaling, trained with $\lambda=0.03$. We do not
ablate logits-only versus values-only injection, alternative $\lambda$
values, per-head gates, or per-layer thresholds. These ablations are
needed to separate the contribution of the sigmoid substrate from the
contribution of the particular gate parameterization.

Finally, H$_2$O and KeyDiff cover two complementary post-hoc signals:
accumulated attention importance and attention-free key similarity. We use a fixed H$_2$O heavy-hitter/recent split and do not tune post-hoc
baseline hyperparameters, so these comparisons should be interpreted as
matched-cache reference points rather than fully optimized SOTA
benchmarking \citep{bui2025trimkv,zeng2024attentiongate,chari2025kvdistill,kim2026fastkvzip,xiao2025duoattention,wan2024d2o}. 

We also report PPL
and compression rates, not wall-clock speed. We did not implement a
production fused kernel for the specific gated sigmoid attention path
studied here, despite existing work on fused attention kernels for
softmax and sigmoid attention \citep{dao2022flashattention,ramapuram2024sigmoid}.
Thus the current results should be read as evidence about cache quality
and learnability, not as an end-to-end inference-speed claim.

\bibliography{sigate}
\bibliographystyle{icml2026}

\clearpage
\onecolumn
\appendix

\section{Additional Validation Sweep Results}
\label{app:tau_sweep}

\noindent
\cref{tab:app_tau_sweep} reports the full validation threshold grid.
Each entry gives decode-window PPL and realized hard-KV compression at
that threshold on the 100-sequence validation set. Bold entries are the
selected operating points under the $\dPPL<0.1$ rule.

\begin{table}[h]
\centering
\small
\caption{Full validation threshold sweep.}
\label{tab:app_tau_sweep}
\setlength{\tabcolsep}{10pt}
\renewcommand{\arraystretch}{1.08}
\begin{tabular}{ccccc}
\toprule
$\tau$ & SM+G RoPE & Sig+G RoPE & SM+G NoPE & Sig+G NoPE \\
\midrule
0.000 & 19.62 / 0.0\% & 19.48 / 0.0\% & 20.24 / 0.0\% & 21.12 / 0.0\% \\
0.005 & 19.63 / 13.4\% & 19.48 / 14.2\% & 20.24 / 6.5\% & 21.11 / 27.3\% \\
0.010 & \textbf{19.69 / 18.7\%} & 19.48 / 14.6\% & \textbf{20.26 / 9.7\%} & 21.11 / 28.4\% \\
0.020 & 19.96 / 25.3\% & 19.48 / 15.6\% & 20.36 / 13.7\% & 21.11 / 29.3\% \\
0.050 & 21.03 / 32.9\% & \textbf{19.48 / 18.8\%} & 21.57 / 21.8\% & \textbf{21.16 / 31.5\%} \\
0.100 & 23.13 / 40.7\% & 26.02 / 24.8\% & 995.48 / 34.6\% & 22.35 / 36.6\% \\
0.150 & 29.71 / 48.4\% & 31.73 / 36.2\% & 1524.68 / 44.4\% & 28.46 / 43.3\% \\
0.200 & 117.85 / 56.6\% & 42.51 / 46.2\% & 2786.99 / 52.5\% & 78.84 / 50.6\% \\
0.300 & 504.08 / 70.0\% & 79.86 / 59.7\% & 2224.26 / 62.6\% & 228.40 / 60.3\% \\
0.400 & 736.59 / 77.9\% & 111.25 / 68.7\% & 2727.09 / 72.7\% & 206.53 / 68.7\% \\
0.500 & 856.37 / 82.7\% & 156.11 / 76.0\% & 3489.68 / 79.0\% & 214.38 / 73.2\% \\
0.600 & 1125.44 / 86.7\% & 215.09 / 81.0\% & 4659.00 / 84.4\% & 270.07 / 75.6\% \\
0.700 & 1368.69 / 88.5\% & 280.60 / 83.7\% & 4585.06 / 87.7\% & 292.23 / 77.9\% \\
0.800 & 1557.72 / 90.0\% & 361.24 / 85.7\% & 5063.09 / 89.5\% & 297.74 / 80.2\% \\
0.900 & 1913.99 / 90.8\% & 451.68 / 87.4\% & 5493.55 / 90.6\% & 309.61 / 81.3\% \\
\bottomrule
\end{tabular}
\end{table}

\section{Qualitative Gate Heatmaps}
\label{app:gate_heatmaps}

This appendix shows qualitative gate heatmaps on five short prompts:
narrative, technical, dialogue, factual, and question answering. Rows
correspond to gate layers and columns correspond to token positions.
Green indicates larger gate values and red indicates smaller gate
values. These figures are not used as quantitative metrics, but they
illustrate that learned gates vary substantially across layers, token
positions, and substrates rather than acting as uniform cache shrinkers.

\begingroup
\setlength{\floatsep}{6pt}
\setlength{\textfloatsep}{6pt}
\setlength{\intextsep}{6pt}
\setlength{\abovecaptionskip}{3pt}
\setlength{\belowcaptionskip}{0pt}

\begin{figure}[h]
\centering
\includegraphics[width=\textwidth]{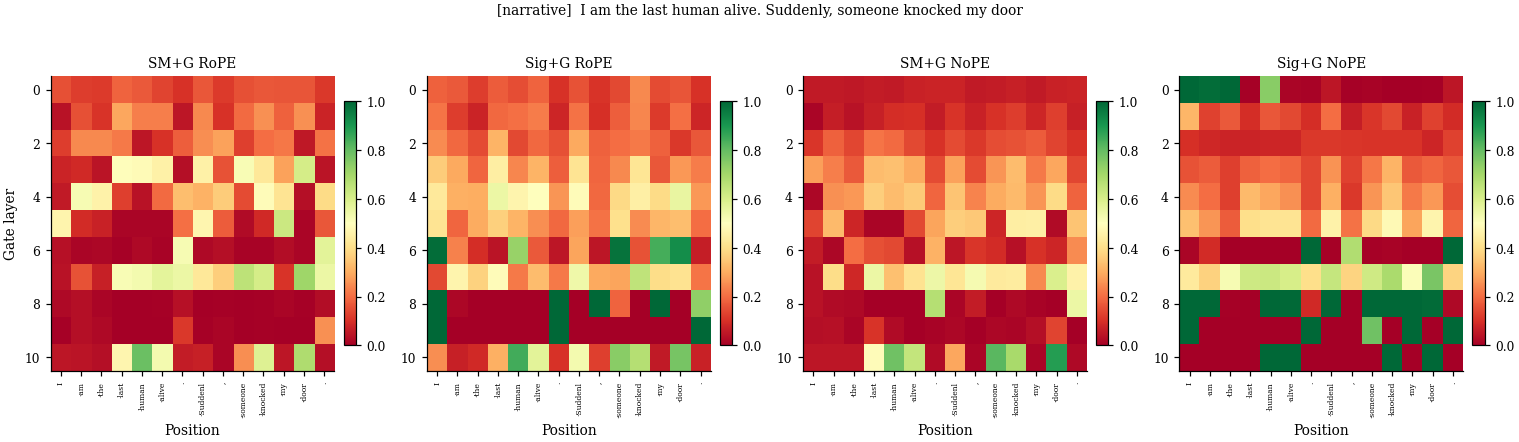}
\caption{Gate heatmap on a narrative prompt.}
\label{fig:app_gate_heatmap_narrative}
\end{figure}

\begin{figure}[h]
\centering
\includegraphics[width=\textwidth]{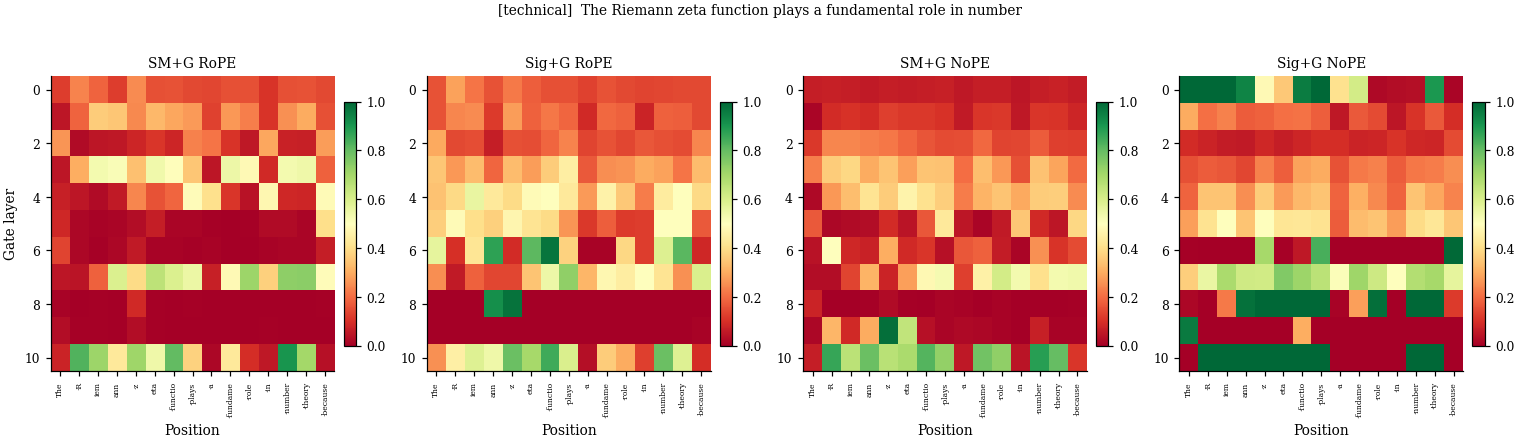}
\caption{Gate heatmap on a technical prompt.}
\label{fig:app_gate_heatmap_technical}
\end{figure}

\begin{figure}[h]
\centering
\includegraphics[width=\textwidth]{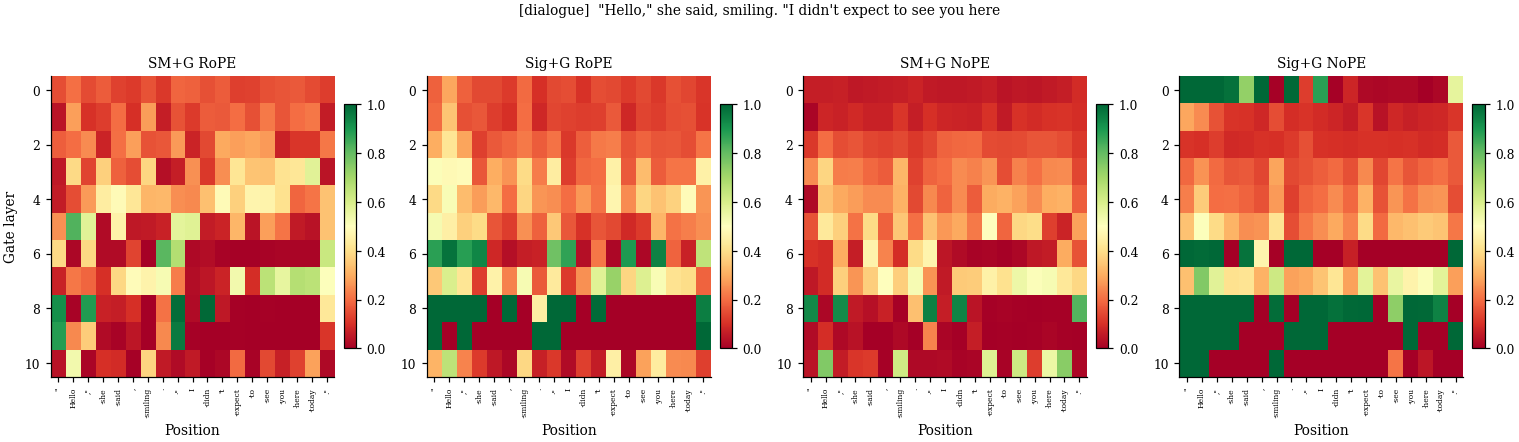}
\caption{Gate heatmap on a dialogue prompt.}
\label{fig:app_gate_heatmap_dialogue}
\end{figure}

\begin{figure}[h]
\centering
\includegraphics[width=\textwidth]{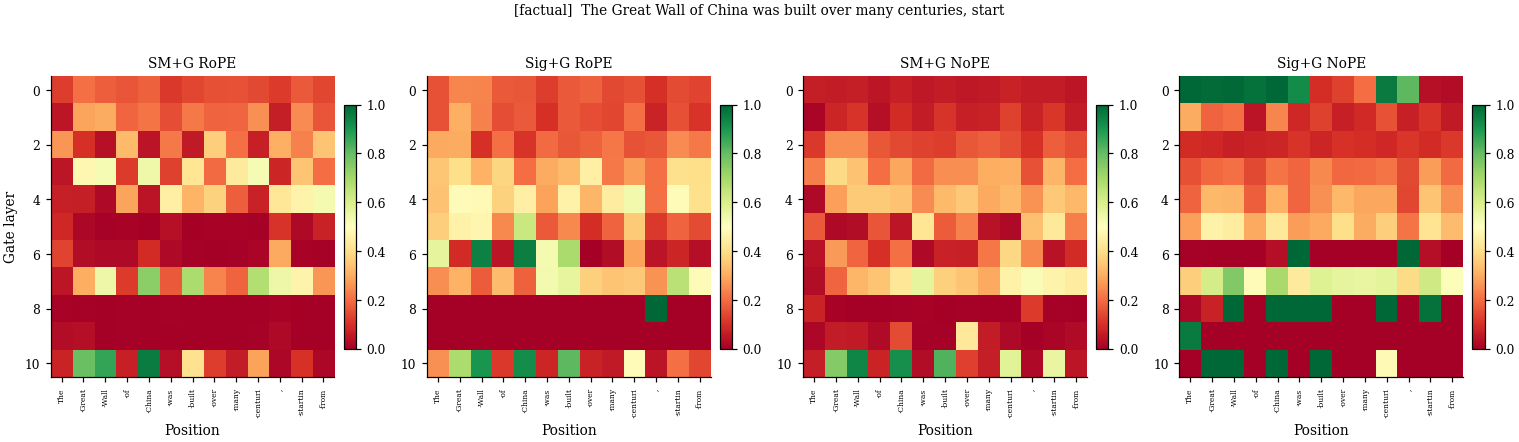}
\caption{Gate heatmap on a factual prompt.}
\label{fig:app_gate_heatmap_factual}
\end{figure}

\begin{figure}[h]
\centering
\includegraphics[width=\textwidth]{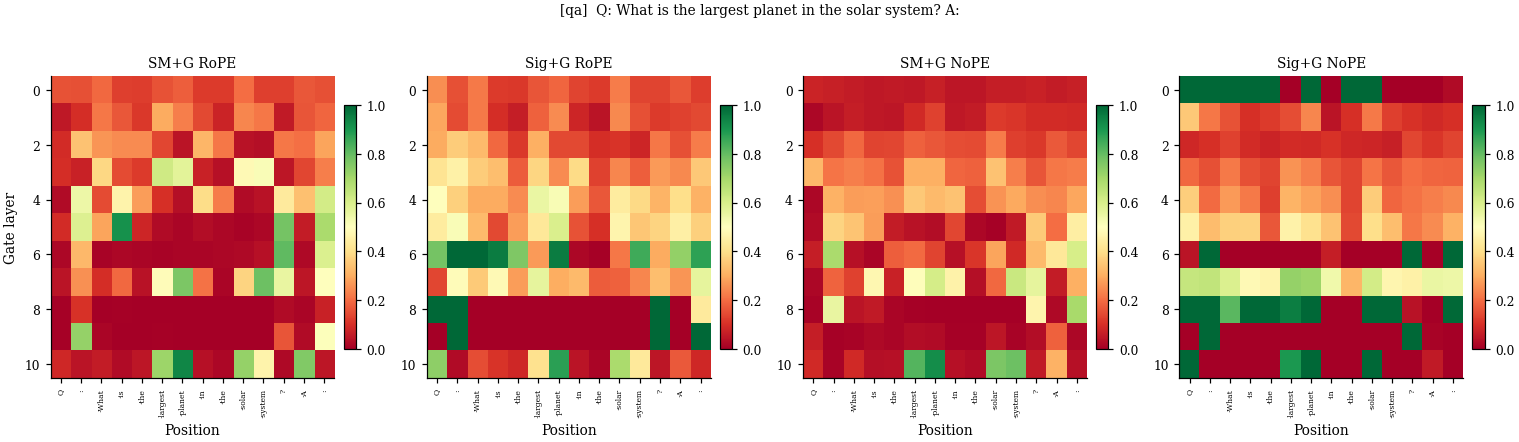}
\caption{Gate heatmap on a question-answering prompt.}
\label{fig:app_gate_heatmap_qa}
\end{figure}

\endgroup

\FloatBarrier

\end{document}